\documentclass{article} 
\usepackage{nips13submit_e,times}
\usepackage{hyperref}
\usepackage{url}
\usepackage{bm}
\usepackage{graphicx}
\usepackage[utf8]{inputenc} 
\usepackage[T1]{fontenc}    
\usepackage{hyperref}       
\usepackage{url}            
\usepackage{booktabs}       
\usepackage{amsfonts}       
\usepackage{nicefrac}       
\usepackage{microtype}      
\usepackage{bm}             
\usepackage{graphicx}       
\usepackage{algorithm}      
\usepackage[noend]{algpseudocode} 
\usepackage{caption}        
\usepackage{array}          
\usepackage{booktabs}       
\usepackage{pbox}           
\usepackage{subcaption}     
\usepackage{multicol}       
\usepackage{tabularx}       
\usepackage{rotating}       

\hypersetup{
    colorlinks = true,
}

\title{Capsule networks for low-data transfer learning}

\author{
Andrew Gritsevskiy* and Maksym Korablyov\\
\texttt{MIT Media Lab} \\
**\texttt{agritsevskiy@gmail.com} \\
}

\nipsfinalcopy 

\begin{document}

\maketitle

\begin{abstract}
 We propose a capsule network-based architecture for generalizing learning to new data with few examples. Using both generative and non-generative capsule networks with intermediate routing, we are able to generalize to new information over 25 times faster than a similar convolutional neural network. We train the networks on the multiMNIST dataset lacking one digit. After the networks reach their maximum accuracy, we inject 1-100 examples of the missing digit into the training set, and measure the number of batches needed to return to a comparable level of accuracy. We then discuss the improvement in low-data transfer learning that capsule networks bring, and propose future directions for capsule research.
\end{abstract}

\section{Introduction}

Unlike the human brain, which can learn from relatively few examples, neural networks have long required tens of thousands of examples to generalize to a new concept. A newborn, having seen a giraffe one or two times in its life, can immediately recognize the animal in any other place. A neural network, however, still requires thousands of giraffe images to have even a comparable level of accuracy. The field of transfer learning focuses on solving this problem, in which we propose a solution using a capsule network.

Capsule networks, first proposed by Sabour et al. \cite{sabour2017dynamic}, are a recently developed type of neural network, in which groups of neurons, called capsules, learn certain properties about an object. Each capsule in a layer tries to predict how to maximize the activations of the higher-level capsules by choosing the amount of information to send to each capsule in the next layer. Deciding where the information is actually sent is determined by the agreement of the capsules’ predictions. Capsules, rather than simply passing along a scalar, maintain a vector of information about the object—where the length of the vector represents the probability that a certain object exists, while the values of the vector represent its instantiation parameters—anything from its orientation to stroke width to color. Each capsule uses a squash nonlinearity, which makes short vectors have a length close to 0 and long vectors have a length close to 1. However, where the capsules send their information is not decided purely by backpropagation. Instead, each layer undergoes several iterations of dynamic routing—the capsules try to predict the output of those in the next layer, and send information based on agreement. It is this procedure that allows capsule-based architectures to so effectively segment a scene into its underlying objects without considering parts of different objects to be spatially related. However, despite the potential present in capsule-based architectures, their accuracy on image segmentation tasks has not been considerably higher than that of their convolutional counterparts. In this paper, we demonstrate that capsule networks—in particular, generative capsule networks—provide a tremendous advantage over convolutional neural networks in the area of transfer learning, where they can learn segment a scene with unknown objects over 100 times faster than before, as well as achieve a significantly higher overall accuracy on an expanded dataset.

\section{Architectures}

We focus on three architectures in our paper---a regular convolutional neural network, a regular capsule network (as described in Sabour et al.), and a generative capsule network. The convolutional neural network consists of three convolutional layers and two fully-connected layers with softmax cross-entropy loss. The regular capsule network contains routing between the Primary Capsule layer and the Digit Capsule layer, as in the original paper. The generative capsule network uses what we call a memo architecture, which consists of convolving the images into the Digit Capsules, applying convolutional reconstruction, and classifying images based on the reconstruction (a detailed description of the setup and implementation can be found in \textbf{Appendix I}).

\section{Methods}
\begin{figure}[h]
  \caption{Three training images from the multiMNIST dataset. The first image contains a 0 overlapping with a 9, the second contains a 2 overlapping with a 5, and the third contains a 4 overlapping with a 0.}
  \centering
    \includegraphics[width=0.25\textwidth]{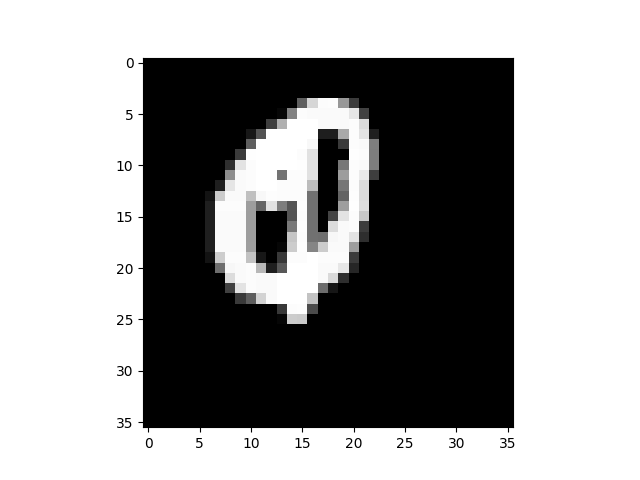}
    \includegraphics[width=0.25\textwidth]{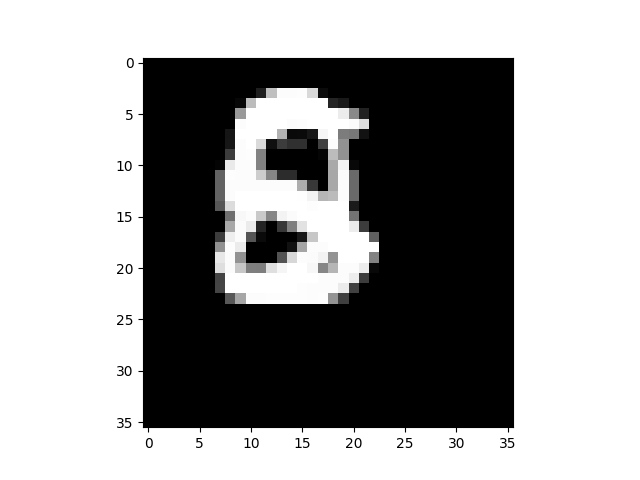}
    \includegraphics[width=0.25\textwidth]{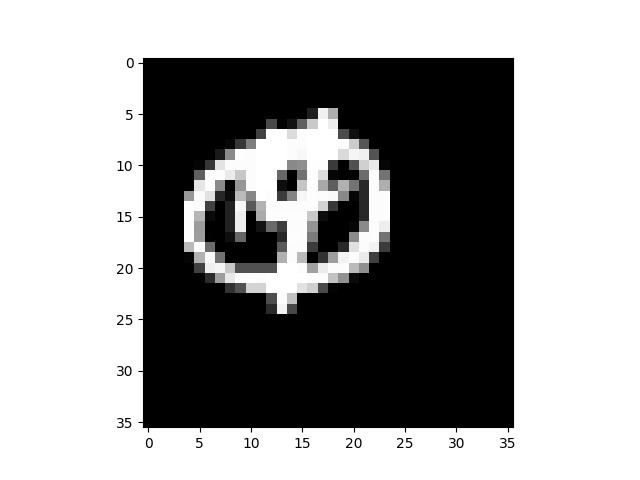}
  
  \label{multiMNIST_example}
\end{figure}

Our experiments were conducted on variations of the multiMNIST dataset, first described in Sabour et al \cite{sabour2017dynamic}. The task is to cleave overlapping MNIST images, such as those in Figure \ref{multiMNIST_example}, into the two constituent digits. The three datasets we used, summarized in Table I, were as follows: first, we had a normal multiMNIST dataset with 200000 training examples of randomly overlapping MNIST images, with each image shifted uniformly randomly between 0 to 8 pixels relative to the other image. This also contained 10000 test examples of overlapping digits shifted by each of 0, 2, 4, 6, and 8 pixels relative to each other. Next, we had the subMMNIST dataset, which is identical in principle to the multiMNIST dataset, except the training data did not contain any examples with a certain single digit. One test set similarly lacked this certain digit; the second test set consisted of examples comprised of all ten digits. Finally, we also had the ldMMNIST dataset, which, instead of completely lacking a digit, contained only a small number of examples of that digit. Specifically, we had eight such ldMMNIST datasets, containing 1, 5, 10, 50, 100, 500, 1000, and 5000 examples, respectively. The test sets were identical to those of the subMMNIST dataset.

\begin{table}[h]
\centering
 \begin{tabular}{|p{2cm} || p{3cm} | p{2.8cm}|  p{2.8cm}|}
 \hline
 Dataset & Training set & Testing sets & Train/test examples \\ [0.5ex] 
 \hline
 multiMNIST & Ten digits & Ten digits shifted from 0 to 8px & 200000 total/10000 per test set \\ 
 subMMNIST & Nine digits & Ten digits shifted from 0 to 8px & 200000 total/10000 per test set \\
 ldMMNIST & Ten digits, but one digit is only sampled from (1/10/50/100/...) examples & Ten digits shifted from 0 to 8px & 200000 total/10000 per test set \\
 \hline
 \end{tabular}
 
\caption{A summary of the three datasets used.}
\label{datasets}
\end{table}

\begin{figure}[h]
  \caption{A sampling of training examples containing a 6 from the ldMMNIST-1 (ld-1) dataset. Note how in each of the images, the example used for the 6 is the same.}
  \centering
    \includegraphics[width=0.17\textwidth]{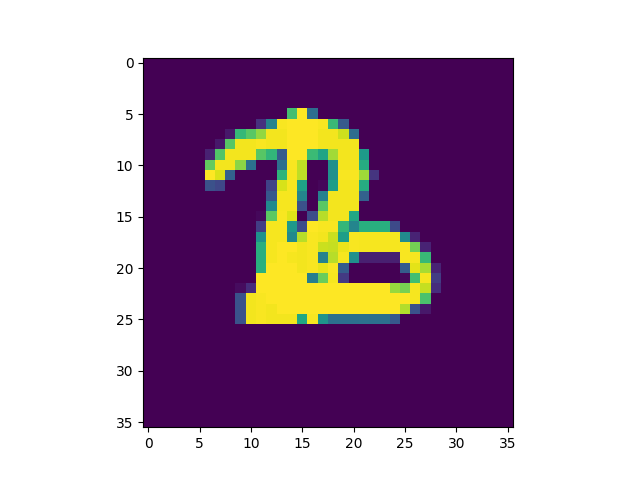}
    \includegraphics[width=0.17\textwidth]{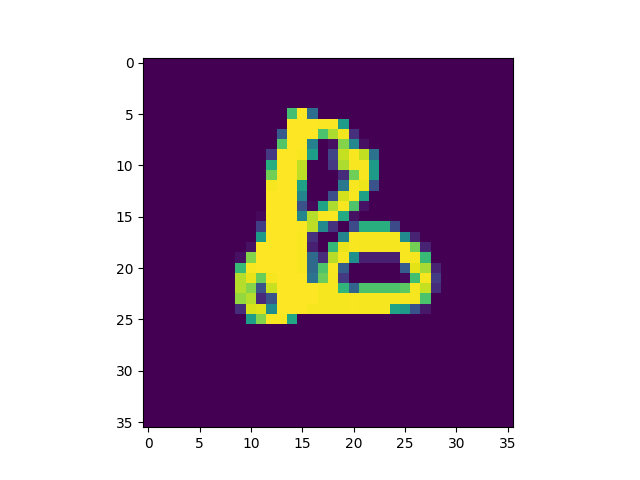}
    \includegraphics[width=0.17\textwidth]{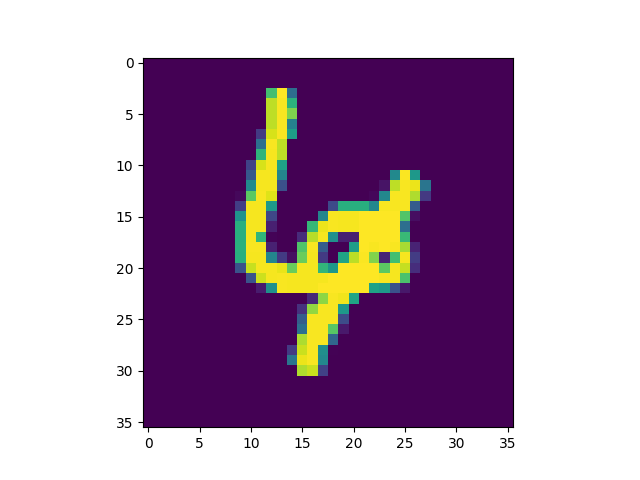}
    \includegraphics[width=0.17\textwidth]{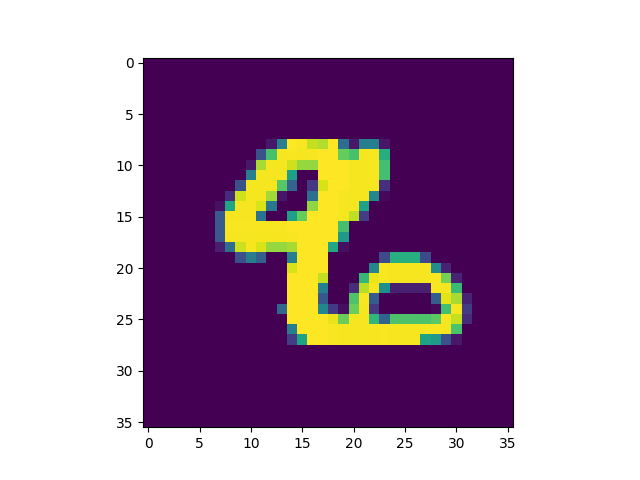}
    \includegraphics[width=0.17\textwidth]{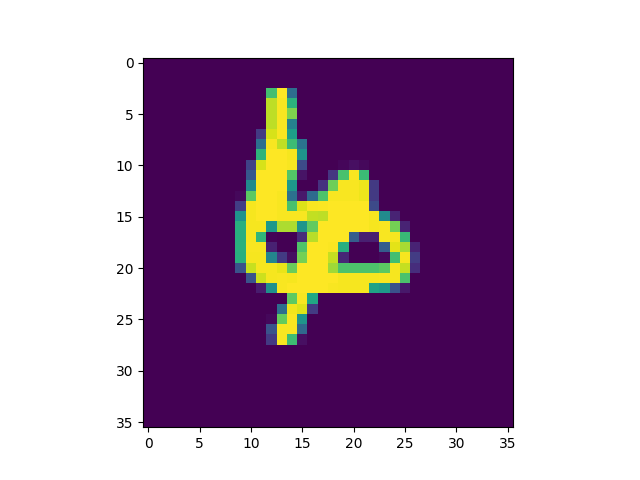}
  
  \label{ld1-example}
\end{figure}

The experimental setup was essentially training the three different architectures on the subMMNIST dataset, and proceeding to inject the missing digit into the training data by switching to either the multiMNIST or the ldMMNIST datasets. Then, we would see how quickly each network could adjust to new data by reaching its pre-injection accuracy. We also analyzed the final accuracy it attained.

\section{Experimental results}

\subsection{Generative capsule network and convolutional network}

For our first comparison, we trained the convolutional network and the generative capsule network for 125,000 batches each on the subMMNIST dataset. Both networks achieved an accuracy of roughly 81\%, with the capsGAN scoring slightly higher on the test set. After the 125,000 iterations, we injected the multiMNIST dataset, which contains all the digits, as well as the ldMMNIST-100 dataset, which contains all the digits but only 100 distinct examples of the tenth digit. It took the convolutional network 2700 iterations to reach its old accuracy, and it achieved less than an 82\% final accuracy with the full injection, and an 88.4\% final accuracy with the ld-100 injection. The generative capsule network adapted to the new data instantaneously (within 100 iterations), and achieved over a 96.3\% peak accuracy with the full injection, and a 97.5\% peak accuracy with the ld-100 injection, as shown in Table \ref{capsGANvsconv@125k}. A plot of the testing accuracy is shown in Figure \ref{capsGANvsconv@125k_testaccuracygraph}, and the raw data is provided in \textbf{Appendix II}.

\begin{table}[h]
    \centering
    \begin{tabular}{l | p{2cm} | p{2cm} | p{2cm} | p{2cm}}
    Architecture              & Iterations to reach initial accuracy & Pre-injection accuracy & Peak accuracy on full test set with full injection & Peak accuracy on full test set with ld-100 injection \\
    \hline
    Convolutional & 2700 & 80.7\% & <82\% & 88.4\% \\
    Generative capsule            & \textbf{<100} & 81.9\% & \textbf{96.3\%} & \textbf{97.5\%} \\
    \end{tabular}
    \caption{The generative capsule architecture is achieves better accuracy at a faster rate when generalizing to new data, when compared to standard convolutional techniques.}
    \label{capsGANvsconv@125k}
\end{table}

Thus, in experiments comparing the speed and effectiveness of adaptation to new data, the generative capsule model significantly outperformed conventional convolutional architectures, improving upon data generalization speed by over a factor of 25, and final accuracy by over 9 percentage points, both on full- and low-data injections.

\begin{figure}[]
  \caption{A plot of the training accuracies of the generative capsule network (capsGAN) and the convolutional network (convnet) from iteration 100,000 to 135,000, with the injections occurring on the 125,000th iteration.}
  \centering
    \includegraphics[width=1.0\textwidth]{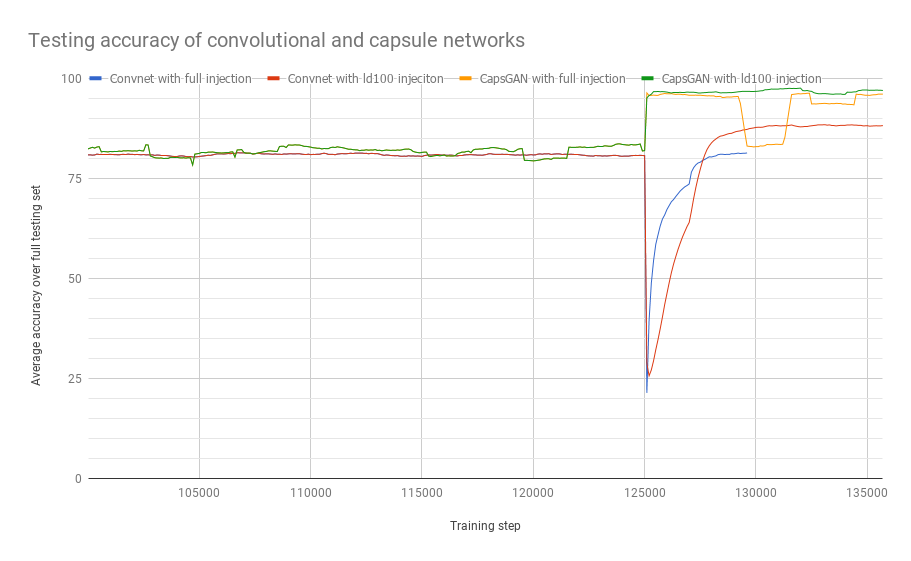}
  
  \label{capsGANvsconv@125k_testaccuracygraph}
\end{figure}

\subsection{Capsule network and convolutional network}

\begin{figure}[]
  \caption{Testing accuracy of a capsule network and a convolutional network, before and after a full-data injection.}
  \centering
    \includegraphics[width=1.0\textwidth]{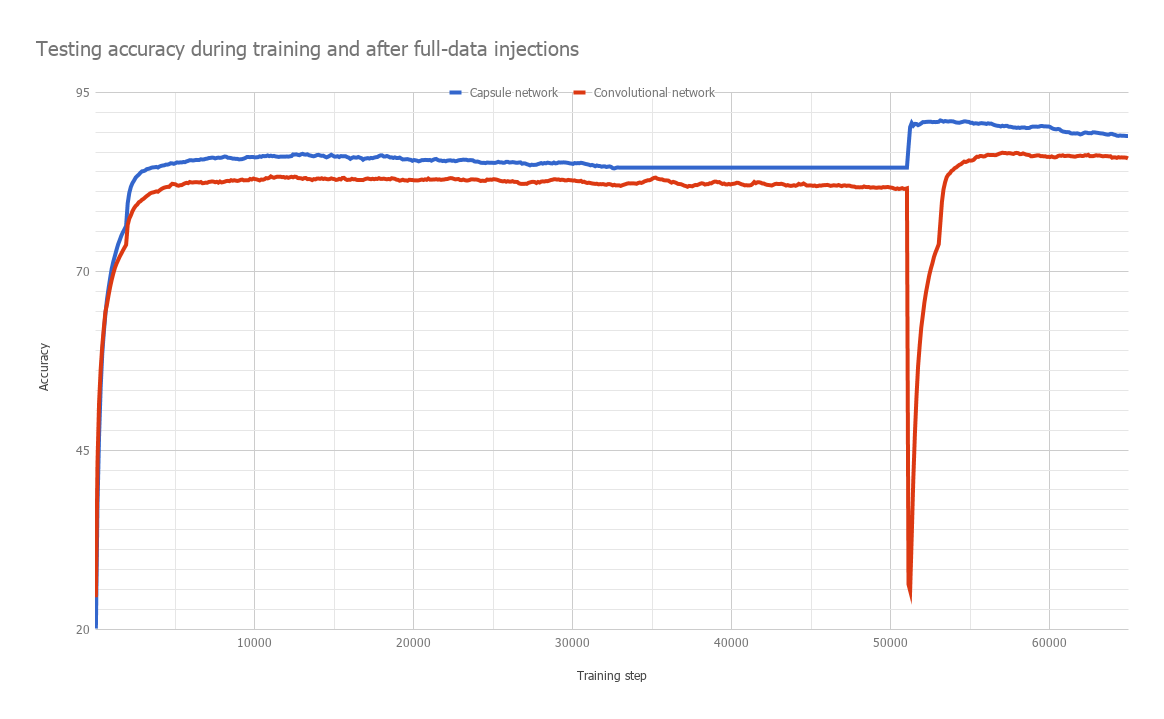}
  
  \label{fig:training-and-injection}
\end{figure}

\begin{sidewaysfigure}[ht]
    \includegraphics[width=\textwidth]{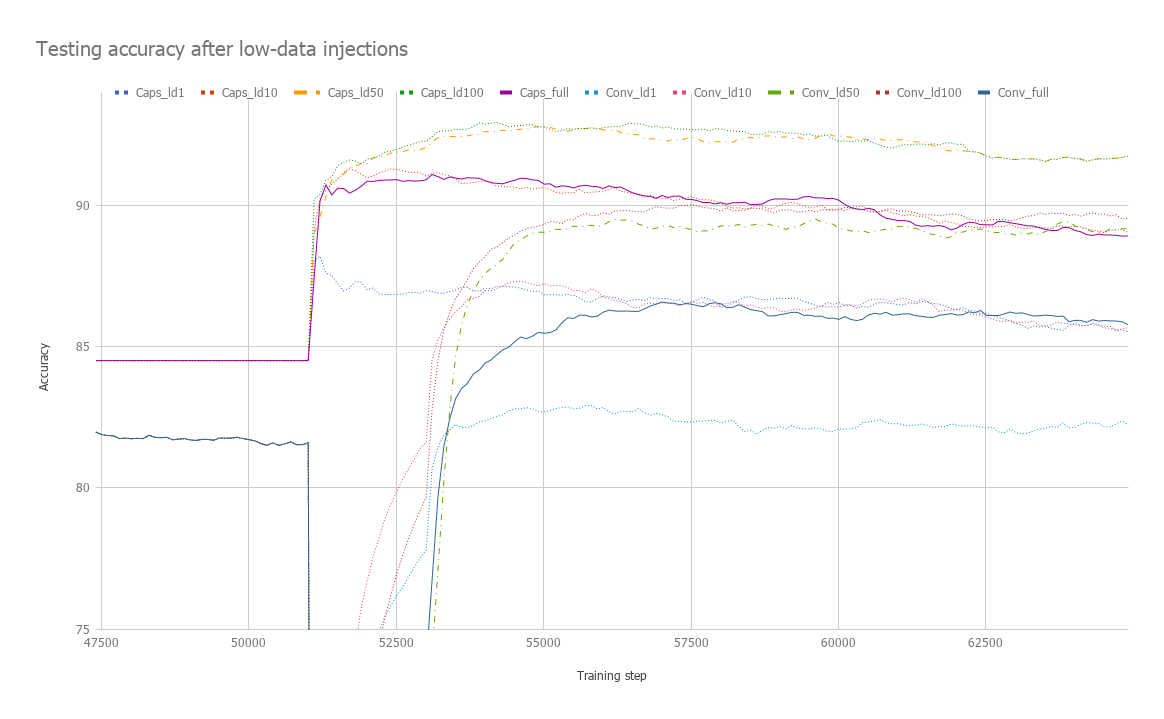}
    \caption{Caption in landscape to a figure in landscape.}
    \label{fig:injection-detail}
\end{sidewaysfigure}

We also wish to show that generative learning is not a requirement for capsule networks to generalize well to new data. In our second experiment, we train a convolutional network and a capsule network for 50,000 iterations each\footnote{In reality, the capsule network was only trained for 33000 iterations, but this should not have a significant effect on the results. We have considered the idea that if a convolutional network trains longer than its capsule counterpart, the L2 regularization may make the weights smaller and therefore less adaptive to new data. However, a capsule injection at 7500 iterations achieved virtually identical results, from which we conclude that this is not a significant issue.} before injecting the full datasets, as well as the ld-1, ld-10, ld-50, and ld-100 test sets. Unlike the generative capsule network, the regular capsule network immediately achieves a higher accuracy while training, reaching 84.5\% before injection, compared to the convolutional network's 81.4\%. After injection, the capsule network also outperforms the convolutional network with a 91.1\% peak accuracy compared to the latter's 86.5\%, as plotted in Figure \ref{fig:training-and-injection}. The most interesting results were perhaps on the low-data injections---shockingly, a capsule network presented with 50 or 100 examples of a missing digit greatly outperformed the same network on the full dataset, by over a percentage point in accuracy. A complete summary of low-data training is described in Table \ref{capsGANvsconv@125k} and Figure \ref{fig:injection-detail}.

\begin{table}[h]
    \centering
    \begin{tabular}{l | p{2cm} | p{2cm} | p{1cm}  | p{1cm}  | p{1cm}  | p{1cm}  | p{1cm}}
    \multicolumn{3}{c}{\phantom{a}} & \multicolumn{4}{|c}{Peak accuracy on dataset after injection} \\
    \hline
    Architecture              & Iterations to reach initial accuracy & Pre-injection accuracy & ld-1 & ld-10 & ld-10 & ld-100 & full \\
    \hline
    Convolutional & 2300 & 81.4\% & 82.9\% & 87.3\% & 89.5\% & 90.0\% & 86.5\%\\
    Capsule            & <100 & 84.5\% & 88.2\% & 91.3\% & 92.8\% & 93.0\% & 91.1\% \\
    \end{tabular}
    \caption{The capsule net improves upon the convolutional network in every test of data generalization.}
    \label{capsGANvsconv@125k}
\end{table}

\section{Discussion}
There are several major results to note about capsule networks. First, capsule networks, both standard and generative, are a significant improvement upon existing convolutional neural networks in the realm of generalizing to new data, steadily reaching improvements of nine percentage points (for generative networks) and three percentage points (for standard networks) over their convolutional counterparts after data injection. One major reason for this may be intrinsic to the structure of the capsule network itself---namely, it consists of ten "pathways" through which information about digits can be routed. If only nine digits are presented, as is the case initially, only nine of the ten pathways may be activated. When a new digit is presented, it does not activate any of the capsules in the existing routing pathways, and is thus sent through the previously unused one. Thus, we predict that capsule networks will have trouble generalizing to more data than channels available in the network, but further testing is required to verify this hypothesis. Furthermore, this may help explain the baffling result of a higher accuracy on a low-data injection (50/100 examples) than on a full-data injection---if there are fewer examples, it may become clearer to the network that this is a separate class of digit. If more examples are presented, it may have trouble classifying it as a new digit, and instead try to group each example into an existing class. Once again, further research is required to find the optimal amount of new data for the best possible generalization. However, it is still acutely intriguing that, even with convolutional networks, and, to a greater extent, with capsule networks, less data in transfer learning may have an unprecendented positive impact.

Another interesting point is the difference between generative capsule networks and regular capsule networks. Although generative capsule networks are both slower and worse at training, achieving a pre-injection accuracy of several percentage points less and taking thousands of iterations longer to achieve it, they nevertheless perform much better after injection---achieving as high as a 97.5\% accuracy on the multiMNIST test set, improving upon the 95.0\% currently achieved by Sabour et al.'s capsule network and the 96.7\% CapsNet-Dropout-2 architecture described in Korablyov \cite{korablyov_2017}. Most remarkable, however, is the result that if all the data is provided to the generative capsule network at once, it cannot attain this accuracy---perhaps, a data injection with new information helps the network better sort its information. Maybe networks are not yet perfect at organizing data on their own, and providing data in a certain order, at intervals, can improve learning. Until, of course, the capsule networks learn to do that themselves.

\section*{Acknowledgements}
The authors would like to thank Dr. Joseph Jacobson and the MIT Media Lab for their incredible guidance and support. Andrew Gritsevskiy would also like to thank Maksym Korablyov for his constant invaluable advice, as well as the the MIT PRIMES program, especially its organizers Slava Gerovitch, Pavel Etingof, Tanya Khovanova, and Srinivas Devadas for this incredible research opportunity; as well as Derik Kauffman.

\nocite{*}
\bibliographystyle{alpha}
\bibliography{lio}

\newcommand{\etalchar}[1]{$^{#1}$}
\begin{thebibliography}{AAB{\etalchar{+}}16}

\bibitem[AAB{\etalchar{+}}16]{abadi2016tensorflow}
Mart{\'\i}n Abadi, Ashish Agarwal, Paul Barham, Eugene Brevdo, Zhifeng Chen,
  Craig Citro, Greg~S Corrado, Andy Davis, Jeffrey Dean, Matthieu Devin, et~al.
\newblock Tensorflow: Large-scale machine learning on heterogeneous distributed
  systems.
\newblock {\em arXiv preprint arXiv:1603.04467}, 2016.

\bibitem[aut]{anonymous}
Anonymous authors.
\newblock Matrix capsules with em routing.

\bibitem[KB14]{kingma2014adam}
Diederik~P Kingma and Jimmy Ba.
\newblock Adam: A method for stochastic optimization.
\newblock {\em arXiv preprint arXiv:1412.6980}, 2014.

\bibitem[Kor17]{korablyov_2017}
Maksym Korablyov.
\newblock Invariance in capsule networks.
\newblock page 1–8, 2017.

\bibitem[SFH17]{sabour2017dynamic}
Sara Sabour, Nicholas Frosst, and Geoffrey~E Hinton.
\newblock Dynamic routing between capsules.
\newblock In {\em Advances in Neural Information Processing Systems}, pages
  3859--3869, 2017.

\end{thebibliography}

\section*{Appendix I: The generative capsule architecture}\label{appendix_i}

The memo architecture begins with a relatively standard capsule network—we take an input image, convolve it up to the digit capsule layer, and then reconstruct the original image. However, we also use the digit capsules for what we call the memo reconstruction. Namely, we reconstruct ten images, each using the maximal activations of a digit capsule. Using these images, each representing a potential image segmentation, we classify the two digits present in the original image. Thus, instead of minimizing just the margin and reconstruction losses of the original digit capsules, we also minimize the margin and reconstruction losses of the memo. This incentivizes our network to learn segmentation reconstructions that are easy to classify later on, which increases the overall reconstruction quality of our network without sacrificing any accuracy.

\begin{figure}[]
  \caption{The encoder network---using two convolutional layers and two capsule layers, it convolves a multiMNIST image up to the DigitCaps layer.}
  \centering
    \includegraphics[width=1.0\textwidth]{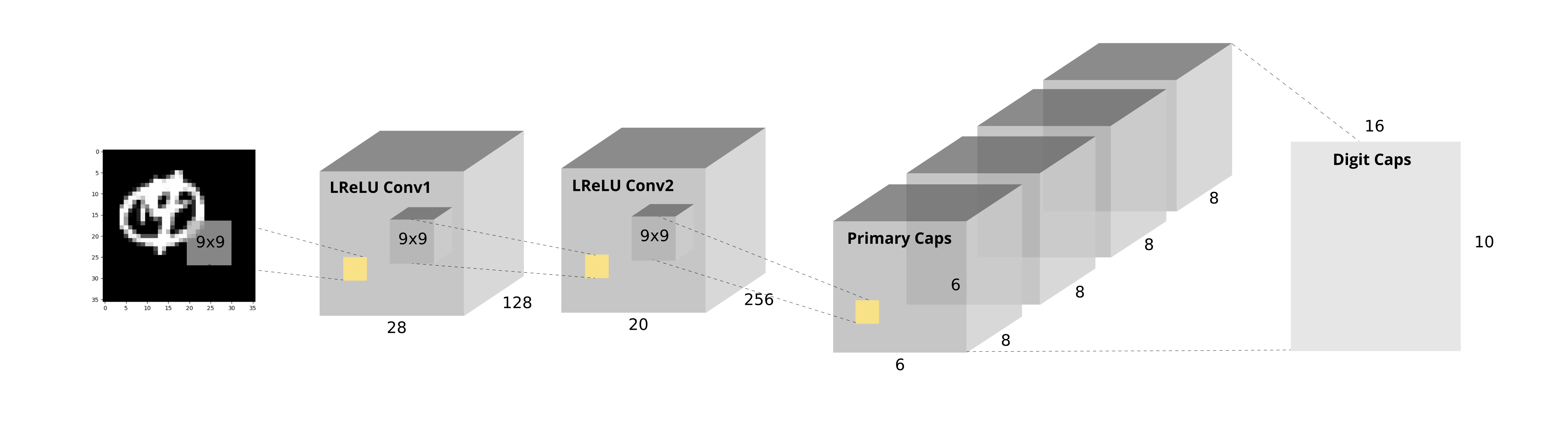}
  
  \label{img_encoder}
\end{figure}

The encoder network, shown in Figure \ref{img_encoder}, is very similar to the encoder network in Sabour et al. \cite{sabour2017dynamic}. The main difference is that for the multiMNIST dataset, we were able to achieve better performance using two convolutions: the first one with a 9x9 kernel with depth 1 and stride 1 using a LeakyReLU activation and 128 filters, and the second one with a 9x9 kernel with depth 128 and stride 1 using a LeakyReLU activation and 256 filters. The alpha values for the LeakyReLU activations were 0.2 and 0.15, respectively. The Conv2 layer was then convolved up to the PrimaryCaps layer using a similar 9x9 kernel convolution with depth 256, but with stride 2 and a filter size of 32*8, using a normal ReLU activation. Finally, the PrimaryCaps layer is routed to the DigitCaps layer. Analogously to the original paper, we only use routing between two consecutive capsule layers: in this case, the PrimaryCaps and DigitCaps layers.

\begin{figure}[]
  \caption{The image decoder network. Using two convolutional and three fully-connected layers, it converts the digit capsules back into a 36x36 image.}
  \centering
    \includegraphics[width=0.9\textwidth]{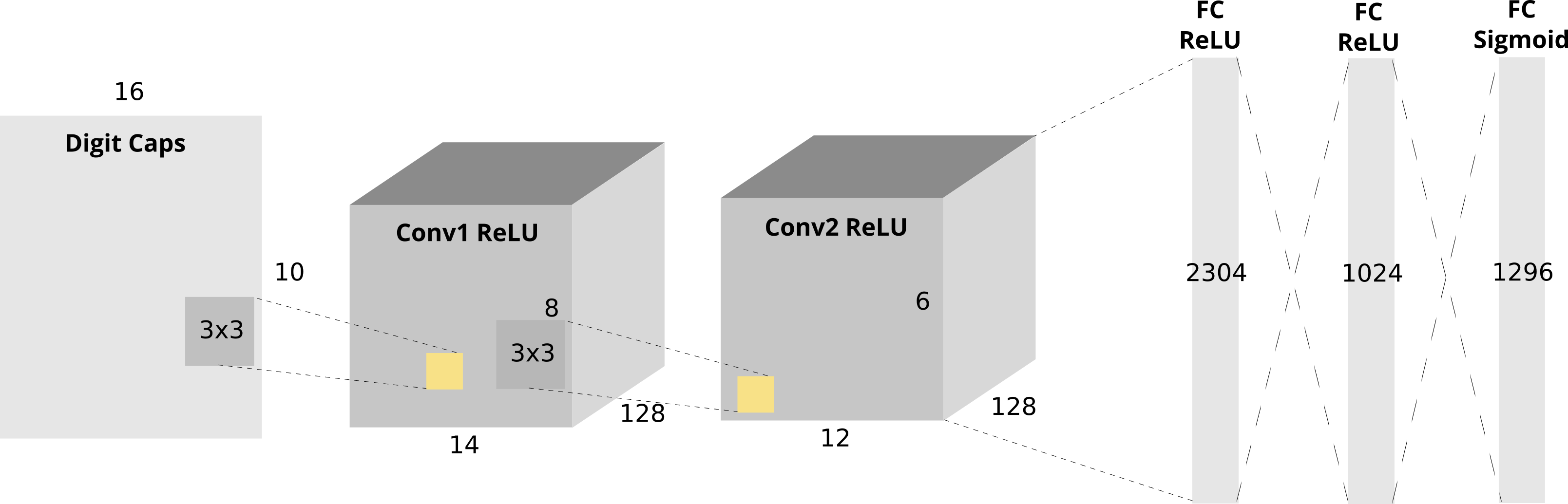}
  
  \label{img_decoder}
\end{figure}

The image decoder network, shown in Figure \ref{img_decoder}, uses two convolutional and three fully-connected layers in order to decode the digit capsules back into the 36x36 image. Both convolutions have 3x3 strides, filters of depth 128, and use simple ReLU activations. The three fully connected layers have 2304, 1024, and 1296 neurons, respectively. Applying the two convolutional layers here has an unexpected benefit---although the accuracy does not increase significantly, we achieve state-of-the-art performance over three times faster than an analogous network containing only fully-connected layers. However, convolutional reconstruction has an issue of instability, which we discuss in the results section.

\begin{figure}[]
  \caption{The memo decoder network, which decodes every capsule activation to see which digits are present.}
  \centering
    \includegraphics[width=0.9\textwidth]{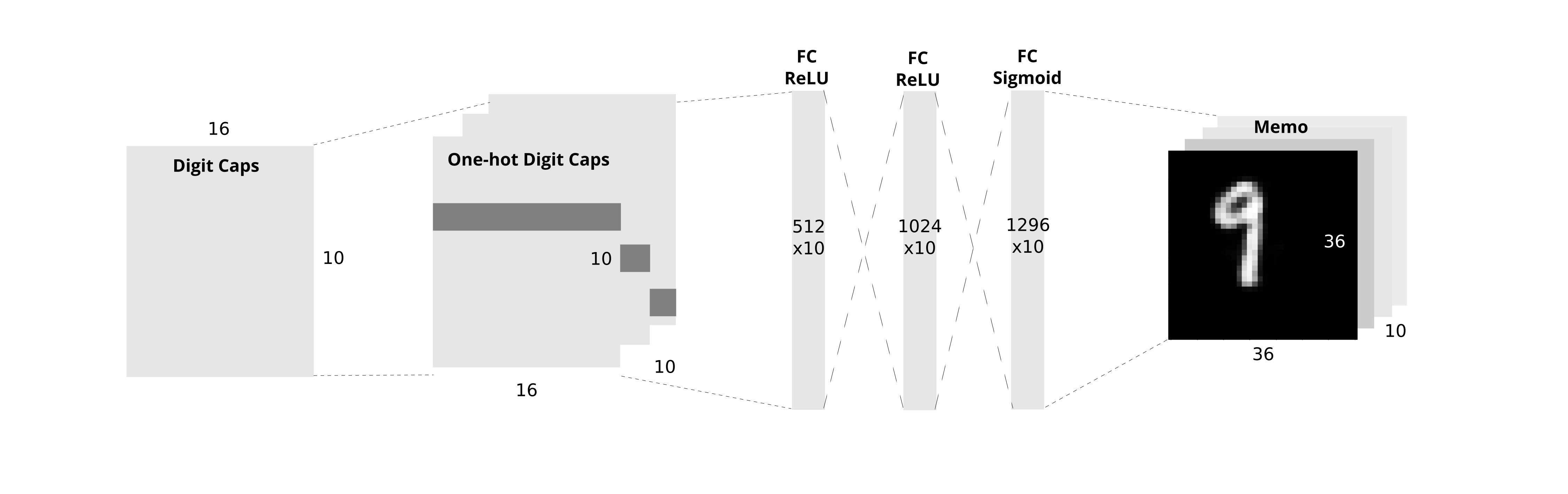}
  
  \label{memo_decoder}
\end{figure}

The memo decoder network, shown in Figure \ref{memo_decoder}, is the fundamental part of the memo architecture. We create ten one-hot encodings of the digit caps, one with each capsule activated. Applying three fully-connected layers for reconstruction, we get a [36x36x10] tensor containing a reconstruction attempt for each digit. The best two of ten are combined into the final reconstructed image.

\begin{figure}[]
  \caption{The final memo encoder, which convolves from the memo back to the digit capsules. Although not shown in the diagram, the depth of the first convolution is 11, in order to see all of the capsule reconstructions, and not just one at a time.}
  \centering
    \includegraphics[width=1.0\textwidth]{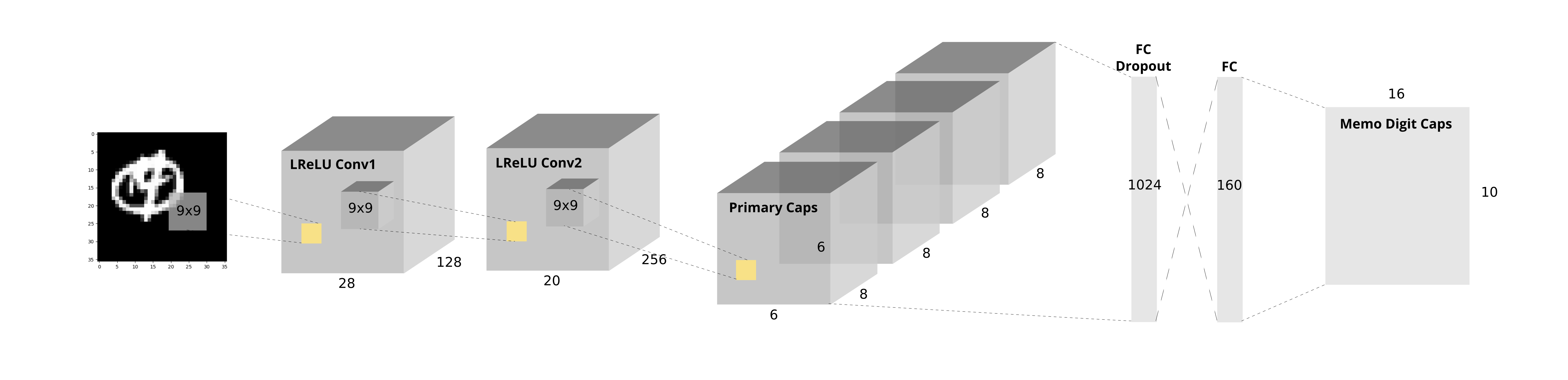}
  
  \label{final_encoding}
\end{figure}

Finally, we reach the last encoding part of the network—the conversion of the memo images to the digit capsules. As shown in Figure \ref{final_encoding}, we start out by applying the same convolutions up to the Primary Caps layer as in the very beginning of the network. However, instead of using the usual algorithm for getting digit capsules from the primary capsule layer, we use a 1024-neuron fully connected layer, followed by a dropout of 0.5, and followed by a 160-neuron fully connected layer. The result of this layer is then reshaped into the [16x10] digit capsules tensor. It is important to note that, since there are no consecutive capsule layers in this encoder, \textit{no routing takes place}! The memo encoder uses capsules exclusively for their activations without optimizing agreement.

\begin{figure}[]
  \caption{The full memo architecture, with two encoders and two decoders. Losses in \textbf{black} are trained with the Training AdamOptimizer, while ones in \textbf{\textcolor{green}{green}} are trained using the Memo AdamOptimizer. As in the original paper, the reconstruction losses are scaled by a factor of 0.005 as to not dominate the error term. Finally, it is important to note that the reconstruction loss acts as a regularizer over the network.}
  \centering
    \includegraphics[width=1.0\textwidth]{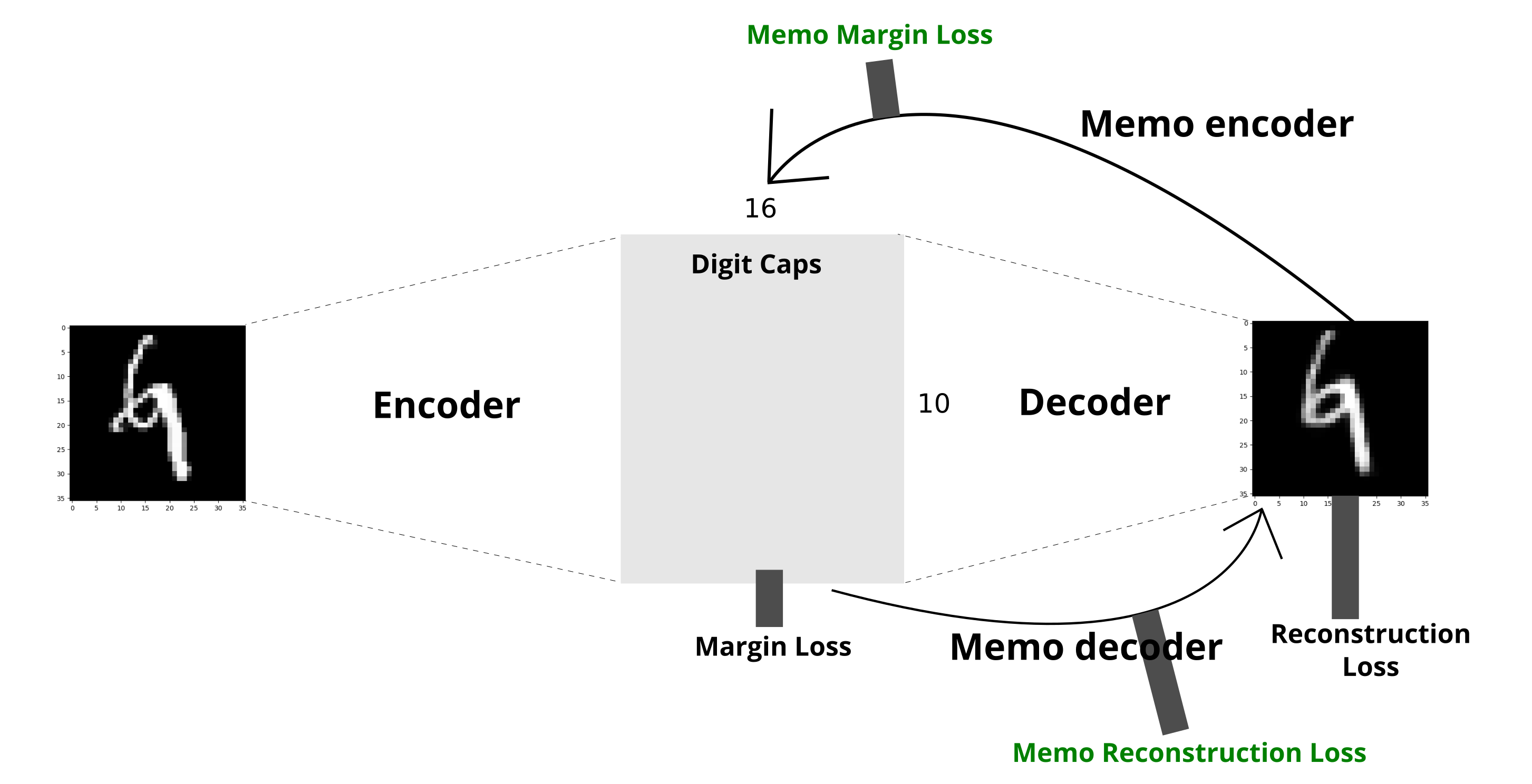}
  
  \label{full_network}
\end{figure}

\subsection{Pre-training} \label{pre-training}
When training our network, we utilize a strategy called pre-training: for the first 1000 iterations or so, only the margin and reconstruction loss of the first encoder-decoder is minimized. After the 1000 iterations, both the main loss and the memo loss are minimized. This allows the network to start out with a reasonable accuracy in the first encoder, which significantly accelerates the improvement rate from the addition of the memo decoder-encoder.

\subsection{Training}
As shown in Figure \ref{full_network}, we use two AdamOptimizers in order to minimize loss \cite{kingma2014adam}. The train margin loss is based on the margin loss described in Sabour et al., as shown in Eq. \ref{margin_loss}.

\begin{equation}
L_k = T_k \ \max(0, m^{+} - ||{\bf v}_k||)^2 + \lambda \ (1-T_k) \ \max(0, ||{\bf v}_k|| - m^{-})^2
\label{margin_loss}
\end{equation}

The reconstruction loss is simply the total squared difference between the reconstructed image and the original. The training AdamOptimizer's loss consists of the margin loss plus the reconstruction loss scaled by a factor of 0.005. The memo AdamOptimizer's loss consists of the margin loss from the reconstruction classification, plus the second reconstruction loss scaled by a factor of 0.005. Both optimizers are trained concurrently, except for the pretraining period described in Section \ref{pre-training}.

\section*{Appendix II: Accuracy data for convolutional and generative capsule networks at 125k-iteration injection}

\fontsize{5}{6}\selectfont

\begin{multicols}{3}
[Listed in order: Step, Accuracy of convnet with full injection, Accuracy of convnet with ld-100 injeciton, Accuracy of capsGAN with full injection, Accuracy of capsGAN with ld-100 injection]
100009,0.80878905,0.80878905,0.82378906,0.82378906\\
100109,0.809023432,0.809023432,0.82589845,0.82589845\\
100209,0.808281252,0.808281252,0.827851556,0.827851556\\
100309,0.80875,0.80875,0.825859384,0.825859384\\
100409,0.810820328,0.810820328,0.828476552,0.828476552\\
100509,0.80996095,0.80996095,0.82941406,0.82941406\\
100609,0.810078116,0.810078116,0.81640626,0.81640626\\
100709,0.809726562,0.809726562,0.817421882,0.817421882\\
100809,0.809999998,0.809999998,0.81683593,0.81683593\\
100909,0.809921876,0.809921876,0.81632814,0.81632814\\
101009,0.810312508,0.810312508,0.817460928,0.817460928\\
101109,0.80980467,0.80980467,0.816875012,0.816875012\\
101209,0.809609382,0.809609382,0.81753908,0.81753908\\
101309,0.80937498,0.80937498,0.817226554,0.817226554\\
101409,0.80910158,0.80910158,0.818359368,0.818359368\\
101509,0.809999992,0.809999992,0.817695316,0.817695316\\
101609,0.810117192,0.810117192,0.819179684,0.819179684\\
101709,0.80980469,0.80980469,0.81886719,0.81886719\\
101809,0.810820308,0.810820308,0.819335928,0.819335928\\
101909,0.81015624,0.81015624,0.818867192,0.818867192\\
102009,0.810195318,0.810195318,0.81855468,0.81855468\\
102109,0.809531252,0.809531252,0.818984372,0.818984372\\
102209,0.81007811,0.81007811,0.81847656,0.81847656\\
102309,0.809492188,0.809492188,0.820351554,0.820351554\\
102409,0.8091797,0.8091797,0.819140642,0.819140642\\
102509,0.809804692,0.809804692,0.818750004,0.818750004\\
102609,0.808906248,0.808906248,0.834140624,0.834140624\\
102709,0.80914061,0.80914061,0.833593746,0.833593746\\
102809,0.808710934,0.808710934,0.806289052,0.806289052\\
102909,0.809062494,0.809062494,0.80417968,0.80417968\\
103009,0.80761719,0.80761719,0.801835922,0.801835922\\
103109,0.80851563,0.80851563,0.801054678,0.801054678\\
103209,0.808242192,0.808242192,0.800937492,0.800937492\\
103309,0.807343732,0.807343732,0.801054694,0.801054694\\
103409,0.8073047,0.8073047,0.799921864,0.799921864\\
103509,0.80742186,0.80742186,0.800195292,0.800195292\\
103609,0.80671874,0.80671874,0.800234366,0.800234366\\
103709,0.805937492,0.805937492,0.802617196,0.802617196\\
103809,0.805117198,0.805117198,0.802773428,0.802773428\\
103909,0.804882812,0.804882812,0.802265632,0.802265632\\
104009,0.80484375,0.80484375,0.80277344,0.80277344\\
104109,0.806054692,0.806054692,0.801640646,0.801640646\\
104209,0.806484366,0.806484366,0.801445316,0.801445316\\
104309,0.805976556,0.805976556,0.801679678,0.801679678\\
104409,0.80480468,0.80480468,0.80093749,0.80093749\\
104509,0.80433593,0.80433593,0.80183596,0.80183596\\
104609,0.804843742,0.804843742,0.801406268,0.801406268\\
104709,0.80359378,0.80359378,0.783789072,0.783789072\\
104809,0.803906256,0.803906256,0.810898438,0.810898438\\
104909,0.804140632,0.804140632,0.811640612,0.811640612\\
105009,0.80488282,0.80488282,0.81265626,0.81265626\\
105109,0.80566407,0.80566407,0.814921866,0.814921866\\
105209,0.806328128,0.806328128,0.814765628,0.814765628\\
105309,0.807070308,0.807070308,0.814257804,0.814257804\\
105409,0.80804687,0.80804687,0.815742206,0.815742206\\
105509,0.807578144,0.807578144,0.81585938,0.81585938\\
105609,0.808984378,0.808984378,0.816015622,0.816015622\\
105709,0.810390638,0.810390638,0.815195318,0.815195318\\
105809,0.811445308,0.811445308,0.814335956,0.814335956\\
105909,0.811328118,0.811328118,0.813906232,0.813906232\\
106009,0.811406228,0.811406228,0.81367188,0.81367188\\
106109,0.811093766,0.811093766,0.813945318,0.813945318\\
106209,0.811601578,0.811601578,0.81457032,0.81457032\\
106309,0.81230468,0.81230468,0.814453118,0.814453118\\
106409,0.811835938,0.811835938,0.8159375,0.8159375\\
106509,0.812734372,0.812734372,0.816523444,0.816523444\\
106609,0.812500002,0.812500002,0.803398452,0.803398452\\
106709,0.814218746,0.814218746,0.82121094,0.82121094\\
106809,0.814062498,0.814062498,0.82144531,0.82144531\\
106909,0.814062498,0.814062498,0.822187508,0.822187508\\
107009,0.813945312,0.813945312,0.815156232,0.815156232\\
107109,0.81332031,0.81332031,0.813125008,0.813125008\\
107209,0.812890626,0.812890626,0.813398448,0.813398448\\
107309,0.812851568,0.812851568,0.81292969,0.81292969\\
107409,0.81113282,0.81113282,0.811289058,0.811289058\\
107509,0.81191405,0.81191405,0.811757842,0.811757842\\
107609,0.812499992,0.812499992,0.81328127,0.81328127\\
107709,0.812539058,0.812539058,0.81449218,0.81449218\\
107809,0.811562478,0.811562478,0.815468748,0.815468748\\
107909,0.811171864,0.811171864,0.81675781,0.81675781\\
108009,0.811406248,0.811406248,0.817421878,0.817421878\\
108109,0.811445318,0.811445318,0.81878906,0.81878906\\
108209,0.81093752,0.81093752,0.818867188,0.818867188\\
108309,0.810312488,0.810312488,0.81828125,0.81828125\\
108409,0.811289068,0.811289068,0.817734382,0.817734382\\
108509,0.810546854,0.810546854,0.8173828,0.8173828\\
108609,0.810625,0.810625,0.82992187,0.82992187\\
108709,0.809882804,0.809882804,0.830429692,0.830429692\\
108809,0.81125,0.81125,0.830312508,0.830312508\\
108909,0.810859374,0.810859374,0.82726564,0.82726564\\
109009,0.81140626,0.81140626,0.833710938,0.833710938\\
109109,0.81183594,0.81183594,0.83339845,0.83339845\\
109209,0.811328128,0.811328128,0.83324218,0.83324218\\
109309,0.811289048,0.811289048,0.83406249,0.83406249\\
109409,0.811835912,0.811835912,0.83347659,0.83347659\\
109509,0.811757798,0.811757798,0.833476572,0.833476572\\
109609,0.810703148,0.810703148,0.832148424,0.832148424\\
109709,0.810078124,0.810078124,0.830312512,0.830312512\\
109809,0.809765622,0.809765622,0.83015626,0.83015626\\
109909,0.810703108,0.810703108,0.829257812,0.829257812\\
110009,0.8094922,0.8094922,0.8283203,0.8283203\\
110109,0.810039062,0.810039062,0.826835922,0.826835922\\
110209,0.81000001,0.81000001,0.826328112,0.826328112\\
110309,0.810585932,0.810585932,0.825468742,0.825468742\\
110409,0.810546886,0.810546886,0.825546874,0.825546874\\
110509,0.81203125,0.81203125,0.823710926,0.823710926\\
110609,0.81308595,0.81308595,0.824804692,0.824804692\\
110709,0.81207032,0.81207032,0.8245703,0.8245703\\
110809,0.81015624,0.81015624,0.824609376,0.824609376\\
110909,0.81019532,0.81019532,0.827226568,0.827226568\\
111009,0.810117182,0.810117182,0.828124992,0.828124992\\
111109,0.809335912,0.809335912,0.827343748,0.827343748\\
111209,0.81089844,0.81089844,0.82910155,0.82910155\\
111309,0.811484372,0.811484372,0.828007808,0.828007808\\
111409,0.81207032,0.81207032,0.827187498,0.827187498\\
111509,0.811796884,0.811796884,0.8260547,0.8260547\\
111609,0.811796872,0.811796872,0.824492196,0.824492196\\
111709,0.81226562,0.81226562,0.823085952,0.823085952\\
111809,0.813124992,0.813124992,0.823007812,0.823007812\\
111909,0.812226552,0.812226552,0.821953112,0.821953112\\
112009,0.813281262,0.813281262,0.821562496,0.821562496\\
112109,0.813437488,0.813437488,0.821249996,0.821249996\\
112209,0.81332032,0.81332032,0.819453122,0.819453122\\
112309,0.813437516,0.813437516,0.820625,0.820625\\
112409,0.812617186,0.812617186,0.820507798,0.820507798\\
112509,0.81285156,0.81285156,0.821640612,0.821640612\\
112609,0.811679684,0.811679684,0.820585948,0.820585948\\
112709,0.8126172,0.8126172,0.821171888,0.821171888\\
112809,0.812500008,0.812500008,0.820898428,0.820898428\\
112909,0.81187501,0.81187501,0.822187496,0.822187496\\
113009,0.812109358,0.812109358,0.819960942,0.819960942\\
113109,0.810664066,0.810664066,0.820507796,0.820507796\\
113209,0.808984378,0.808984378,0.819101552,0.819101552\\
113309,0.807734368,0.807734368,0.820117188,0.820117188\\
113409,0.808671884,0.808671884,0.821054692,0.821054692\\
113509,0.808281238,0.808281238,0.820546874,0.820546874\\
113609,0.807304704,0.807304704,0.82007813,0.82007813\\
113709,0.8070703,0.8070703,0.82058593,0.82058593\\
113809,0.80664064,0.80664064,0.82023439,0.82023439\\
113909,0.807109372,0.807109372,0.82121093,0.82121093\\
114009,0.805742188,0.805742188,0.820507792,0.820507792\\
114109,0.80542969,0.80542969,0.821562502,0.821562502\\
114209,0.805742188,0.805742188,0.822968758,0.822968758\\
114309,0.80546876,0.80546876,0.823281258,0.823281258\\
114409,0.806718752,0.806718752,0.82347656,0.82347656\\
114509,0.805976556,0.805976556,0.82183592,0.82183592\\
114609,0.80636719,0.80636719,0.81750001,0.81750001\\
114709,0.805664042,0.805664042,0.815898412,0.815898412\\
114809,0.806093748,0.806093748,0.81476564,0.81476564\\
114909,0.805742182,0.805742182,0.814023436,0.814023436\\
115009,0.805039072,0.805039072,0.815156242,0.815156242\\
115109,0.8070703,0.8070703,0.815781252,0.815781252\\
115209,0.808007802,0.808007802,0.81582032,0.81582032\\
115309,0.80890626,0.80890626,0.806015622,0.806015622\\
115409,0.808359372,0.808359372,0.804687488,0.804687488\\
115509,0.80871095,0.80871095,0.806171888,0.806171888\\
115609,0.809492192,0.809492192,0.807109372,0.807109372\\
115709,0.810000006,0.810000006,0.80753906,0.80753906\\
115809,0.808867158,0.808867158,0.806601576,0.806601576\\
115909,0.808398438,0.808398438,0.80746093,0.80746093\\
116009,0.808593748,0.808593748,0.808632812,0.808632812\\
116109,0.80808593,0.80808593,0.80597653,0.80597653\\
116209,0.808125008,0.808125008,0.806523444,0.806523444\\
116309,0.808320302,0.808320302,0.805976552,0.805976552\\
116409,0.807812492,0.807812492,0.806015622,0.806015622\\
116509,0.806640626,0.806640626,0.807265612,0.807265612\\
116609,0.80644533,0.80644533,0.812343752,0.812343752\\
116709,0.80750002,0.80750002,0.813749998,0.813749998\\
116809,0.807343752,0.807343752,0.815976572,0.815976572\\
116909,0.808671882,0.808671882,0.815703132,0.815703132\\
117009,0.809179682,0.809179682,0.817460932,0.817460932\\
117109,0.809414062,0.809414062,0.817968738,0.817968738\\
117209,0.809570308,0.809570308,0.81437501,0.81437501\\
117309,0.809101558,0.809101558,0.821054682,0.821054682\\
117409,0.808945314,0.808945314,0.823046878,0.823046878\\
117509,0.808242174,0.808242174,0.823085938,0.823085938\\
117609,0.808437508,0.808437508,0.8236328,0.8236328\\
117709,0.808007816,0.808007816,0.823554664,0.823554664\\
117809,0.808945328,0.808945328,0.824921868,0.824921868\\
117909,0.80992188,0.80992188,0.824960924,0.824960924\\
118009,0.811640632,0.811640632,0.82539063,0.82539063\\
118109,0.811718742,0.811718742,0.82703124,0.82703124\\
118209,0.81042969,0.81042969,0.82722656,0.82722656\\
118309,0.810234382,0.810234382,0.82671875,0.82671875\\
118409,0.81011718,0.81011718,0.82585939,0.82585939\\
118509,0.810078124,0.810078124,0.824843758,0.824843758\\
118609,0.810507822,0.810507822,0.823554668,0.823554668\\
118709,0.809999972,0.809999972,0.82359376,0.82359376\\
118809,0.810312488,0.810312488,0.821757808,0.821757808\\
118909,0.80976562,0.80976562,0.817773456,0.817773456\\
119009,0.80925779,0.80925779,0.817265636,0.817265636\\
119109,0.8089453,0.8089453,0.816796852,0.816796852\\
119209,0.808203122,0.808203122,0.81914062,0.81914062\\
119309,0.809023432,0.809023432,0.822773446,0.822773446\\
119409,0.808281268,0.808281268,0.824257802,0.824257802\\
119509,0.80902344,0.80902344,0.824570312,0.824570312\\
119609,0.80867188,0.80867188,0.795351568,0.795351568\\
119709,0.808007804,0.808007804,0.795195314,0.795195314\\
119809,0.80882814,0.80882814,0.7948828,0.7948828\\
119909,0.80949219,0.80949219,0.79449219,0.79449219\\
120009,0.808710932,0.808710932,0.79398438,0.79398438\\
120109,0.809023448,0.809023448,0.794492186,0.794492186\\
120209,0.810703108,0.810703108,0.795468762,0.795468762\\
120309,0.811406232,0.811406232,0.79625,0.79625\\
120409,0.810507824,0.810507824,0.79765626,0.79765626\\
120509,0.81160156,0.81160156,0.798867202,0.798867202\\
120609,0.811445312,0.811445312,0.799140618,0.799140618\\
120709,0.811718728,0.811718728,0.799101564,0.799101564\\
120809,0.809921882,0.809921882,0.797617192,0.797617192\\
120909,0.8107422,0.8107422,0.800859372,0.800859372\\
121009,0.81124999,0.81124999,0.800742188,0.800742188\\
121109,0.81148438,0.81148438,0.800781248,0.800781248\\
121209,0.81121095,0.81121095,0.80054687,0.80054687\\
121309,0.809843762,0.809843762,0.80097657,0.80097657\\
121409,0.810507828,0.810507828,0.80078125,0.80078125\\
121509,0.81042968,0.81042968,0.800234372,0.800234372\\
121609,0.81027343,0.81027343,0.828085924,0.828085924\\
121709,0.810546874,0.810546874,0.82777342,0.82777342\\
121809,0.810117184,0.810117184,0.82777343,0.82777343\\
121909,0.809726578,0.809726578,0.827734388,0.827734388\\
122009,0.809765628,0.809765628,0.828085954,0.828085954\\
122109,0.809023454,0.809023454,0.828789062,0.828789062\\
122209,0.808203118,0.808203118,0.827734364,0.827734364\\
122309,0.80789062,0.80789062,0.828476558,0.828476558\\
122409,0.806835932,0.806835932,0.827109372,0.827109372\\
122509,0.806367196,0.806367196,0.827421876,0.827421876\\
122609,0.806093752,0.806093752,0.827695308,0.827695308\\
122709,0.805820312,0.805820312,0.827539064,0.827539064\\
122809,0.807500008,0.807500008,0.82890625,0.82890625\\
122909,0.80660158,0.80660158,0.8303906,0.8303906\\
123009,0.80679687,0.80679687,0.829765622,0.829765622\\
123109,0.805898428,0.805898428,0.829687502,0.829687502\\
123209,0.806523454,0.806523454,0.83152342,0.83152342\\
123309,0.806054688,0.806054688,0.831132822,0.831132822\\
123409,0.806679692,0.806679692,0.83035156,0.83035156\\
123509,0.807070328,0.807070328,0.830312508,0.830312508\\
123609,0.8078125,0.8078125,0.832460952,0.832460952\\
123709,0.807617216,0.807617216,0.835312494,0.835312494\\
123809,0.807109366,0.807109366,0.836289062,0.836289062\\
123909,0.80570312,0.80570312,0.836249998,0.836249998\\
124009,0.805546888,0.805546888,0.834843758,0.834843758\\
124109,0.805781258,0.805781258,0.83429688,0.83429688\\
124209,0.805703142,0.805703142,0.833554688,0.833554688\\
124309,0.805976572,0.805976572,0.833242208,0.833242208\\
124409,0.80691406,0.80691406,0.834960938,0.834960938\\
124509,0.80730468,0.80730468,0.83375,0.83375\\
124609,0.807578122,0.807578122,0.833906244,0.833906244\\
124709,0.80734376,0.80734376,0.83472657,0.83472657\\
124809,0.807812516,0.807812516,0.836093752,0.836093752\\
124909,0.807343738,0.807343738,0.818671868,0.818671868\\
125009,0.806992192,0.806992192,0.81957031,0.81957031\\
125109,0.21484375,0.28828125,0.9635416667,0.951171875\\
125209,0.39609375,0.257421876,0.9570312667,0.95703125\\
125309,0.48854168,0.272135418,0.9574652967,0.96028643\\
125409,0.544726572,0.29492187,0.95800782,0.96679687\\
125509,0.58578124,0.322499998,0.9575521033,0.96679685\\
125609,0.609244802,0.347786452,0.9574652967,0.96712242\\
125709,0.631584834,0.375223214,0.95982145,0.966517875\\
125809,0.64882812,0.404394534,0.96093749,0.96679687\\
125909,0.65902777,0.43524305,0.9620949067,0.96636287\\
126009,0.672421878,0.462421874,0.9614583533,0.965625\\
126109,0.682244318,0.489346584,0.9612926,0.96448862\\
126209,0.691861968,0.514713528,0.9616970533,0.963378905\\
126309,0.69759616,0.53647836,0.9606369867,0.96469353\\
126409,0.705078116,0.555189738,0.9598214533,0.96456472\\
126509,0.711979152,0.57302084,0.9604166733,0.96458333\\
126609,0.718945314,0.588916022,0.96020508,0.96496583\\
126709,0.724126852,0.60390625,0.9592524433,0.964499085\\
126809,0.72903646,0.616883672,0.95956309,0.96473523\\
126909,0.732319066,0.629564154,0.9587445067,0.9655633\\
127009,0.736328108,0.640156242,0.9576823,0.965332025\\
127109,0.76585938,0.66824217,0.9576822867,0.96513675\\
127209,0.776874982,0.699804694,0.9578125333,0.96513672\\
127309,0.7840625,0.727578122,0.9579427167,0.96464845\\
127409,0.788710948,0.7520703,0.9575521,0.96347657\\
127509,0.79074218,0.77351564,0.9578125,0.9633789\\
127609,0.795156252,0.792968746,0.9571614467,0.9638672\\
127709,0.797460932,0.808945308,0.9567708333,0.9642578\\
127809,0.80082029,0.82207031,0.9562500167,0.964648425\\
127909,0.804140632,0.831328116,0.9563802,0.96533203\\
128009,0.80367188,0.838710934,0.9561197867,0.96523438\\
128109,0.805156256,0.844296868,0.9563150967,0.965625025\\
128209,0.8063672,0.84855468,0.9541666667,0.966308595\\
128309,0.80953124,0.852656238,0.9546223867,0.96494141\\
128409,0.810429684,0.855664072,0.95488282,0.96367185\\
128509,0.81039063,0.856718744,0.95208334,0.9638672\\
128609,0.809804682,0.85851562,0.95247392,0.96396485\\
128709,0.810195318,0.86058594,0.95345052,0.96406252\\
128809,0.809999992,0.86218749,0.9531900833,0.96445315\\
128909,0.811874996,0.862656242,0.9533203133,0.96425782\\
129009,0.811523416,0.86496095,0.9541666667,0.96484375\\
129109,0.812304702,0.866953108,0.9544270633,0.96552735\\
129209,0.813320332,0.868046884,0.9547525967,0.9663086\\
129309,0.812265628,0.86886718,0.9371744667,0.9667969\\
129409,0.812851564,0.870859372,0.9013020667,0.96748045\\
129509,0.81292969,0.87156249,0.8650390833,0.967285155\\
129609,0.813749988,0.872499998,0.8309244833,0.96728515\\
129709,,0.874453148,0.8300781033,0.96728515\\
129809,,0.87531248,0.8297526067,0.96718748\\
129909,,0.876171852,0.82871095,0.96718748\\
130009,,0.877617182,0.82962238,0.96738283\\
130109,,0.877773436,0.8294270833,0.96787107\\
130209,,0.877695312,0.8313151,0.96865237\\
130309,,0.87804688,0.8309896,0.96904295\\
130409,,0.87941406,0.83190105,0.97128907\\
130509,,0.880820314,0.8350260133,0.97158203\\
130609,,0.88160156,0.834375,0.97216797\\
130709,,0.88121093,0.8346354167,0.97324218\\
130809,,0.88160155,0.83470055,0.97285157\\
130909,,0.88234375,0.8356119967,0.97304685\\
131009,,0.881562492,0.8352213367,0.97304687\\
131109,,0.881093744,0.8348958367,0.97392578\\
131209,,0.881601538,0.8353515667,0.97353515\\
131309,,0.881601556,0.8531249933,0.97421877\\
131409,,0.881757818,0.88919272,0.97441405\\
131509,,0.882617188,0.9250651,0.974999995\\
131609,,0.883515612,0.9594401133,0.974511745\\
131709,,0.881796884,0.96015624,0.9748047\\
131809,,0.88058593,0.9608724,0.974511745\\
131909,,0.87945314,0.96171875,0.974999995\\
132009,,0.878867192,0.9610677,0.97558595\\
132109,,0.87910157,0.9617187333,0.9693359\\
132209,,0.879882806,0.9617187533,0.96884767\\
132309,,0.879726552,0.9627604133,0.9696289\\
132409,,0.880859372,0.9628255167,0.96806638\\
132509,,0.881562508,0.9359375167,0.96748045\\
132609,,0.88226562,0.93626303,0.964160145\\
132709,,0.88289063,0.9359375333,0.962597675\\
132809,,0.883945312,0.9365885467,0.96162112\\
132909,,0.88375,0.9368489667,0.96142577\\
133009,,0.883945316,0.9372395667,0.9618164\\
133109,,0.884140628,0.9374349133,0.96181643\\
133209,,0.883125,0.9369140533,0.96142578\\
133309,,0.883671878,0.9362630233,0.96083985\\
133409,,0.882304692,0.9370442633,0.960351555\\
133509,,0.88210937,0.9367187333,0.96035155\\
133609,,0.88125,0.9374349,0.9609375\\
133709,,0.881992192,0.9372396,0.96083985\\
133809,,0.882499992,0.9373047133,0.96103515\\
133909,,0.88265623,0.93561198,0.959765645\\
134009,,0.882656236,0.9360677,0.959863305\\
134109,,0.883476542,0.9349609367,0.96542968\\
134209,,0.883359368,0.9348307333,0.96513673\\
134309,,0.883906252,0.9347005333,0.96533202\\
134409,,0.883203122,0.9337890533,0.9661133\\
134509,,0.88316407,0.96041668,0.9663086\\
134609,,0.882539068,0.9597005133,0.96923828\\
134709,,0.88269532,0.9598307167,0.969824195\\
134809,,0.881445312,0.9587239667,0.9709961\\
134909,,0.8814453,0.9578124833,0.970703125\\
135009,,0.881796868,0.9578776167,0.97070315\\
135109,,0.88058595,0.9573567667,0.97001955\\
135209,,0.88089844,0.9583333533,0.970214845\\
135309,,0.88128906,0.9589843667,0.97011718\\
135409,,0.881523444,0.95917968,0.970605475\\
135509,,0.881250008,0.9606119933,0.97031252\\
135609,,0.88136718,0.9606770667,0.97021485\\
135709,,0.88195312,0.9606771,0.9696289\\
\end{multicols}

\end{document}